%% file: main.tex
\documentclass[conference]{IEEEtran}
\IEEEoverridecommandlockouts
\usepackage{cite}
\usepackage{amsmath,amssymb,amsfonts}
\usepackage{algorithmic}
\usepackage{graphicx}
\usepackage{textcomp}
\usepackage{booktabs}
\usepackage{multirow}
\usepackage{url}
\usepackage[colorlinks]{hyperref}
\usepackage{xcolor}
\def\BibTeX{{\rm B\kern-.05em{\sc i\kern-.025em b}\kern-.08em
    T\kern-.1667em\lower.7ex\hbox{E}\kern-.125emX}}

\begin{document}

\title{GGMotion: Group Graph Dynamics-Kinematics Network for Human Motion Prediction}

\author{\IEEEauthorblockN{1\textsuperscript{st} Shuaijin Wan{*}}
\IEEEauthorblockA{\textit{dept. School of Computer Science and Engineering} \\
\textit{Nanjing University of Science and Technology}\\
Nanjing, China \\
inkcat@njust.edu.cn}
}

\maketitle

\begin{abstract}
Human motion is a continuous physical process in 3D space, governed by complex dynamic and kinematic constraints. Existing methods typically represent the human pose as an abstract graph structure, neglecting the intrinsic physical dependencies between joints, which increases learning difficulty and makes the model prone to generating unrealistic motions. In this paper, we propose GGMotion, a group graph dynamics-kinematics network that models human topology in groups to better leverage dynamics and kinematics priors. To preserve the geometric equivariance in 3D space, we propose a novel radial field for the graph network that captures more comprehensive spatio-temporal dependencies by aggregating joint features through spatial and temporal edges. Inter-group and intra-group interaction modules are employed to capture the dependencies of joints at different scales. Combined with equivariant multilayer perceptrons (MLP), joint position features are updated in each group through parallelized dynamics-kinematics propagation to improve physical plausibility. Meanwhile, we introduce an auxiliary loss to supervise motion priors during training. Extensive experiments on three standard benchmarks, including Human3.6M, CMU-Mocap, and 3DPW, demonstrate the effectiveness and superiority of our approach, achieving a significant performance margin in short-term motion prediction. The code is available at \url{https://github.com/inkcat520/GGMotion.git}.
\end{abstract}


\section{Introduction}
\begin{figure}
    \centering
    \includegraphics[width=1\linewidth]{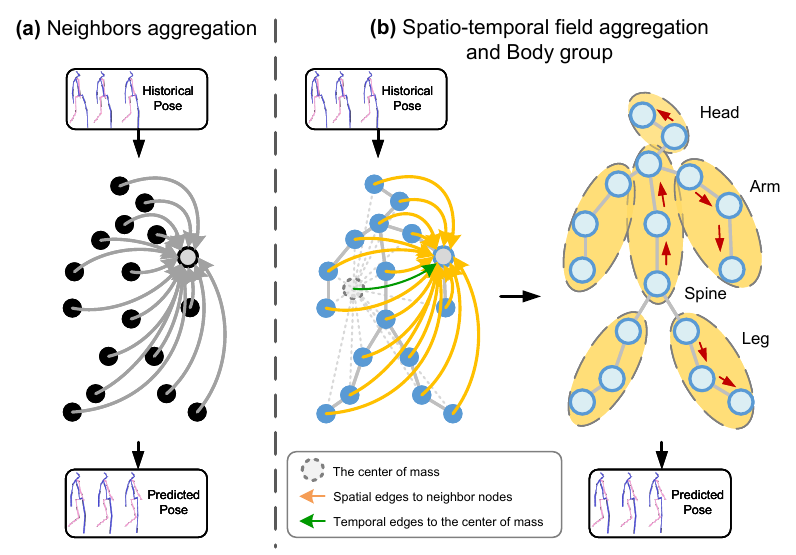}
    \caption{Previous methods perform prediction by aggregating features from neighboring nodes. In contrast, our approach constructs a spatio-temporal radial field composed of edges to aggregate geometric features and synthesize motion forces. Then the human body is partitioned into multiple groups (e.g., spine, legs, arms, and head), and joint positions are updated through both dynamics and kinematics principles. The red arrows indicate the direction of information propagation within each group.}
    \vspace{-3mm}
    \label{fig1}
\end{figure}
Human motion prediction refers to predicting future human skeletal motion poses based on observed motion sequences, which is an important topic in computer vision and plays a significant role in practical applications such as autonomous driving~\cite{autodrivechen20203d, autodriveliang2020learning}, human-computer interaction~\cite{intergui2018teaching, humanrobot}, and animation production~\cite{animationholden2016deep}. The human skeleton can be regarded as a rigid body system composed of multiple joints, and 3D human motion consists of submotions performed by different body parts. Additionally, human motion is a physical process with complex spatio-temporal dependencies in 3D geometric space, which makes it challenging to model explicitly by deterministic algorithms.

Deep learning has emerged as the predominant approach to human motion prediction tasks. Due to Recurrent Neural Networks (RNN)~\cite{ACRNN, ERD, GAILRNN, DAE, PVRED, Structural-RNN, RES-GRU, SPLRNN, LongRNN} struggling to capture spatial information, existing studies predominantly focus on Transformer~\cite{history-transformer, potr, Multi-transformer, ST-transformer,humanmac, auxformer, realistic-transformer}, and Graph Convolution Networks (GCN)~\cite{DMGCN, GAGCN, MSRGCN, STSGCN, TrajGCN, PGBIGGCN, SPGSNGCN, KSOFGCN, Skeleton-Aware} to model the spatial structure of human poses. The iterative prediction paradigm~\cite{SPLRNN, jointpropagation-transformer} in a chain-like dependency tends to accumulate errors along the chain, degrading the accuracy of the extremity joint predictions. Meanwhile, \cite{SPGSNGCN, Skeleton-Aware} have attempted to split the body into upper and lower parts to capture the motion characteristics of different body parts. Further, \cite{DPNet} collaboratively models joints with similar dynamic information to optimize implicit correlations. Despite achieving promising results on benchmark datasets, these approaches remain at the human topology modeling level without delving into the inherent dynamics and kinematics constraints of human motion, for example, joints perform rigid-body motion around their parent joints, which makes them prone to generating physically implausible poses. On the other hand, since human motion occurs in 3D geometric space, it is essential to maintain geometric equivariance under Euclidean transformations (i.e., translation, rotation, and reflection) to achieve better generalization with fewer parameters, as demonstrated in~\cite{EQGNN, Eqmotion}. 

In this paper, we propose a Group Graph Dynamics-Kinematics Network, which models the human skeleton as a specialized graph preserving equivariant geometric information and applies parametric dynamics-kinematics algorithms through the group strategy for more accurate and physically plausible human motion prediction. Previous methods that aggregate neighboring node features to update node trajectory features often result in the loss of critical geometric information. 
While equivariant methods\cite{EQGNN, GMN, Eqmotion} employ radial fields to aggregate only spatial features, they struggle to capture temporal trajectory information, limiting their effectiveness in long-term motion prediction. Moreover, they require additional invariant features for each node to represent spatial edges. In contrast, we propose a simple spatio-temporal radial field shown in Fig. \ref{fig1}, which updates position features based on geometric edge features between nodes, including only Euclidean distances and the number of traversed hops. In addition, our method further learns the motion features of each node relative to the centroid, allowing the model to capture richer temporal dependencies and improve continuous motion prediction. More importantly, we propose learnable scaling factors into the radial field to adaptively adjust the influence of spatio-temporal edge features, enabling the model to capture the heterogeneity of different joints for better prediction performance.

To incorporate dynamics and kinematics into the network, we improve the equivariant message passing function~\cite{GMN} by integrating the self-attention mechanism of the Transformer~\cite{Attention-transformer}. Specifically, we replace the conventional softmax-based attention computation with a covariance matrix-based operation to preserve the equivariance of the MLP. Instead of \cite{SPGSNGCN, Skeleton-Aware}, we divide the human skeleton into multiple groups specifically for the dynamics and kinematics modeling shown in Fig. \ref{fig1}. Compared to full-body modeling, the grouping strategy is more beneficial for learning the differences between various body parts. Based on the equivariant MLP, the inter-group and intra-group interaction modules are designed to capture motion force dependencies at different scales. Finally, rigid-body dynamics and kinematics propagation are employed in each group to update the joint position features in parallel, contributing to more accurate and physically plausible motion predictions. An auxiliary loss based on joint length is introduced to impose prior constraints on human motion.

Extensive evaluations are conducted on three benchmark datasets, including Human3.6M, CMU-Mocap, and 3DPW for both short-term and long-term prediction tasks. Experimental results demonstrate that our lightweight approach achieves outstanding performance.

\subsubsection*{\normalfont In summary, our contributions are as follows}
\begin{itemize}
\item{We propose a group graph dynamics-kinematics network that models human topology in groups for parametric dynamics-kinematics algorithms to improve physical plausibility and accuracy of motion prediction.}
\item{We introduce a novel spatio-temporal radial field that adaptively aggregates node features based on geometric edge information, effectively enhancing the modeling of complex spatio-temporal dependencies.}
\item{We propose an equivariant MLP integrated with the self-attention mechanism to ensure geometric equivariance of features in Euclidean space, which is employed in both inter-group and intra-group interaction modules, as well as in dynamics and kinematics propagation.}
\item{Extensive experiments demonstrate the effectiveness of each module and confirm the superior performance of our approach on three benchmark datasets.}
\end{itemize}

\section{Related Work}
\subsection{Human Motion Prediction}
The early research on human motion prediction mainly focuses on traditional probabilistic models~\cite{Probabilitymodel1, Probabilitymodel2, Probabilitymodel3}, which directly impose prior knowledge about human motion. Nevertheless, these methods struggle to generate continuous and reasonable human poses. Subsequently, the field of human motion prediction entered an era driven by deep learning. Since human motion sequences exhibit similarities to time series data, Recurrent Neural Networks (RNN)~\cite{ACRNN, ERD, GAILRNN, DAE, PVRED, Structural-RNN, RES-GRU, LongRNN, SPLRNN} are employed to capture temporal dependencies. To address the inherent cumulative error problem in RNN-based networks, researchers have employed various strategies, e.g., residual structures~\cite{RES-GRU, PVRED}, generative adversarial training~\cite{GANRNN}, and error adaptation mechanisms~\cite{ACRNN}. The spatiotemporal characteristics of human motion motivated researchers~\cite{S2SCNN, TrajCNN, STACNN} to employ sequence-to-sequence convolutional neural networks (CNN) to capture complex spatiotemporal dependencies, which predict sequences from sequences and significantly mitigate the cumulative error between predicted frames. The attention mechanism~\cite{history-transformer, potr, Multi-transformer, ST-transformer,humanmac, auxformer, realistic-transformer} is also highly effective in modeling spatiotemporal sequences. Dai \textit{et al.}~\cite{kdformer} introduced manually computed kinematic features as an additional input. However, reliance on hand-crafted features restricts the ability to learn robust and expressive motion representations.

Graph Convolutional Networks (GCN)~\cite{DMGCN, GAGCN, MSRGCN, STSGCN, TrajGCN, PGBIGGCN, SPGSNGCN, KSOFGCN, Skeleton-Aware} are well-suited for nongrid data and have been widely used to model the spatial dependencies of human joints. Mao \textit{et al.}~\cite{TrajGCN} represented the human body as a fully connected, learnable GCN that adaptively captures the topological structure of the human skeleton. Ma \textit{et al.}~\cite{PGBIGGCN} leverages both spatial and temporal graph convolutions to capture spatio-temporal dependencies. Li \textit{et al.}~\cite{SPGSNGCN} introduced a skeleton-separated graph scattering network, which decomposes pose features using band-pass graph filters to extract rich temporal information. Ding \textit{et al.}~\cite{KSOFGCN} adopts TopK-pooling to achieve the optimal fusion of temporal and spatial features.

Although previous works focus on modeling complex spatial dependencies among joints, they neglect the inherent geometric constraints. The nonlinear transformations introduced by activation functions often break the geometric property, forcing the model to introduce additional parameters to approximate these Euclidean transformations. This explains why siMLPe~\cite{simlp} achieves competitive prediction performance using only linear layers.

\subsection{Equivariant Graph Neural Networks}
To simulate complex rigid-body physical systems, it is crucial to preserve geometric constraints in Euclidean space. Satorras \textit{et al.}~\cite{EQGNN} introduces a simple and effective equivariant message-passing mechanism on graphs, ensuring that the output graph undergoes the same transformation as the input graph under Euclidean transformations. Huang \textit{et al.}~\cite{GMN} models graphical data using three simple dynamic models incorporating stronger rigid-motion priors. However, both methods are restricted to single-step state transitions and lack the capacity for continuous sequence-to-sequence prediction. Xu \textit{et al.}~\cite{Eqmotion} apply equivariance to Euclidean transformations in human motion prediction and achieve remarkable results. Yu \textit{et al.}~\cite{Pose-Transformed} propose a pose-transformed equivariant network which performs equivariant prediction by applying learned pose transformations to uniquely normalized points.

These works show that preserving the equivariance in the rigid body motion contributes to both the compactness and the robustness. However, they aggregate node features solely based on radial fields from neighbors, which limits their capacity to capture the temporal trajectories and fails to model the complete human kinematic topology.

GGMotion introduces a spatio-temporal radial field to aggregate motion forces, and models the human kinematic topology for future motion prediction through a grouping strategy based on dynamics and kinematics.

\section{Methodology}
\subsection{Problem Formulation}
We model human motion in the trajectory space of each joint to facilitate the prediction of arbitrary-length future motion sequences. Assume that the historical motion of the \(i^{th}\) joint is \(\mathbf{X}_i=[\mathbf{x}_{i}^{(1)},\mathbf{x}_{i}^{(2)},\cdots,\mathbf{x}_{i}^{(T_h)}]\in\mathbb{R}^{D\times T_h}\), where \(T_h\) are the past timestamps and \(D\) is
the coordinate space dimension (here \(D=3\)). Similarly, the future \(i^{th}\) joint with \(T_f\) frames is defined as \(\mathbf{Y}_i=[\mathbf{y}_{i}^{(1)},\mathbf{y}_{i}^{(2)},\cdots,\mathbf{y}_{i}^{(T_f)}]\in\mathbb{R}^{D\times T_f}\). Then, the past motion sequence of the human body is \(\mathbf{X}_{1:N}=[\mathbf{X}_{1}, \mathbf{X}_{2},\cdots,\mathbf{X}_{N}]\in\mathbb{R}^{N\times D\times T_{h}}\), and the future motion poses are \(\mathbf{Y}_{1:N}=[\mathbf{Y}_{1},\mathbf{Y}_{2},\cdots,\mathbf{Y}_{N}]\in\mathbb{R}^{N\times D\times T_{f}}\), where \(N\) represents the total number of human joints. Our network \(\mathcal{F}_{predict}(\cdot)\) needs to predict future pose sequences \(\widehat{\mathbf{Y}}\) as close to the ground truth as possible, given the observed value \(\mathbf{X}\). 

According to the skeleton topology, the human body is divided into \(S\)  groups, each consisting of \(k\) joints, as shown in Fig. \ref{fig1}. We denote the \(i^{th}\) group \(\mathcal{G}_{i}=\{n_{i1}, n_{i2},\cdots,n_{ik}\}\), where \(n_{ij}\) is the global index of the \(j^{th}\) joint in group \(\mathcal{G}_{i}\) at the body scale. \(\mathcal{N}_{i}\) and \(\mathcal{P}_{i}\) respectively represent the sets of neighboring nodes and parent nodes of the joint \(i\).

Our network satisfies equivariance under Euclidean transformations that consist of translation, rotation, and reflection, which is defined as:
\[\mathrm{R}\widehat{\mathbf{Y}}+\mathrm{t}=\mathcal{F}_{predict}(\mathrm{R}\mathbf{X}+\mathrm{t}),\quad\forall\mathrm{R}\in\mathcal{O}(D).\]
Here \(D=3\), \(\mathcal{O}(3)\) stands for the 3D orthogonal group\cite{SE3}, \(\mathrm{R}\in\mathbb{R}^{D\times D}\) is the rotation (or reflection) matrix, and \(\mathrm{t}\in\mathbb{R}^{D}\) is the translation vector.

\subsection{Dynamics-Kinematics Analysis}
\label{DK}
\begin{figure}
    \centering
    \includegraphics[width=1\linewidth]{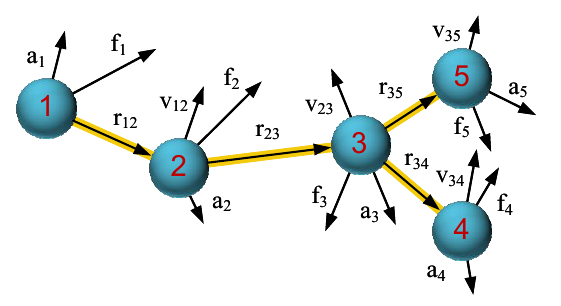}
    \caption{Dynamics and kinematics analysis of the arm group.}
     \vspace{-3mm}
    \label{fig2}
\end{figure}

The key to our method lies in how to effectively model the propagation patterns of human body dynamics and kinematics, while maintaining computational flexibility. As shown in Fig. \ref{fig2}, we demonstrate a simplified dynamic model of the human arm, which is a multistage hinge system composed of interconnected joints. Due to the complexity of the system, we first analyze the rigid-body dynamics~\cite{GMN} only for a two-stage motion system, which is expressed as follows:
\begin{align}
\boldsymbol{a}_{j}=\boldsymbol{a}_{i}+\boldsymbol{\alpha}_{ij}\times\boldsymbol{r}_{ij}+\boldsymbol{\omega}_{ij}\times\boldsymbol{v}_{ij}, \quad i\in\mathcal{P}_{j},
\label{eq:aNext}
\end{align}
where \(a_{j}\in\mathbb{R}^{D\times C}\) denotes the linear acceleration of the parent joint of joint \(j\), while \(\boldsymbol{r}_{ij}\) and \(\boldsymbol{v}_{ij}\) denote the position difference and linear velocity difference from joint \(i\) to joint \(j\). \(\boldsymbol{\alpha}_{ij}=\frac{\boldsymbol{r}_{ij}\times(\boldsymbol{f}_{j}-\boldsymbol{a}_{i})}{\|\boldsymbol{r}_{ij}\|^{2}}\) and \(\boldsymbol{\omega}_{ij}=\frac{\boldsymbol{r}_{ij}\times\boldsymbol{v}_{ij}}{\|\boldsymbol{r}_{ij}\|^2}\) respectively represent the angular velocity and angular acceleration of the joint \(j\) relative to the joint \(i\), and \(\times\) indicates the cross product of vectors. This formulation reveals the propagation relationship of the acceleration between two adjacent joints. However, when extended to multistage joint motion, the dynamics algorithm becomes more complex and difficult to apply directly. Moreover, this iterative propagation process will also lead to the accumulation of prediction errors in joint positions. 

From the above analysis, it is evident that the acceleration of each joint can be determined by three fundamental physical quantities: velocity difference 
\(\boldsymbol{v}\), position difference \(\boldsymbol{r}\), and force \(\boldsymbol{f}\). This relationship further inspires the design of our parallelized dynamics and kinematics propagation combined with a parametric network, which improves computational efficiency while enabling more accurate motion predictions.

\subsection{Equivariant MLP}
To preserve the equivariance of geometric features during the nonlinear transformations of MLPs, we improve the equivariant message passing function proposed in \cite{GMN} by integrating a self-attention mechanism. This enhancement enables the network to better capture the dependencies among variables and strengthens the representational capacity of the MLP. The formulation is as follows:
\[\phi_{eq}(\mathbf{Z}^{n}):=\mathbf{Z}_\mathrm{v}^{n}\mathrm{MLP}(\frac{\boldsymbol{\Sigma}^{n}}{||\boldsymbol{\Sigma}^{n}||_{2,row}}),\quad\boldsymbol{\Sigma}^{n}=\mathbf{Z}_\mathrm{q}^{n\top}\mathbf{Z}_\mathrm{k}^{n},\]
where \(\mathbf{Z}^{n}\) is the concatenation of \(n\) different variables \(\mathbf{Z}\in\mathbb{R}^{D\times C}\) in the last new dimension, and \(\boldsymbol{\Sigma}^{n}\in\mathbb{R}^{n\times n}\) is the covariance matrix of variables in \(\mathbf{Z}^{n}\). \(\boldsymbol{\Sigma}^{n}\in\mathbb{R}^{n\times n}\) is the covariance matrix of the \(D\)-dimensional variables after a linear transformation, and \(||\cdot||_{2, row}\) is \(L_2\)-norm along the row dimension, similar to the attention weight computation accompanied by a softmax function in the Transformer. 

The input \(\mathbf{Z}\) is mapped to \(\mathbf{Z}_\mathrm{q}\in\mathbb{R}^{D\times C}\),\(\mathbf{Z}_\mathbf{k}\in\mathbb{R}^{D\times C}\),\(\mathbf{Z}_\mathbf{v}\in\mathbb{R}^{D\times C}\) through three linear layers \(\mathbf{W}_\mathrm{q}\in\mathbb{R}^{D\times C}\),\(\mathbf{W}_\mathrm{k}\in\mathbb{R}^{D\times C}\),\(\mathbf{W}_\mathrm{v}\in\mathbb{R}^{D\times C}\). After that, the covariance matrix \(\boldsymbol{\Sigma}^{n}\) of the \(n\) variables was nonlinearly transformed in MLP, and \(\mathbf{Z}_\mathbf{v}\) is used to restore the data to the \(D\)-dimensional coordinate space through matrix multiplication. The final output is produced by a linear layer. Since the rotation (or reflection) matrix \(\mathrm{R}\) satisfies \(\mathrm{R}^{\top}\mathrm{R}=\mathrm{I}\), after subtracting the centroid of the pose sequence to eliminate the global translation vector, we can easily conclude that \(\phi_{eq}(\cdot)\) exhibits equivariance.

\begin{figure*}
    \centering
    \includegraphics[width=1\linewidth]{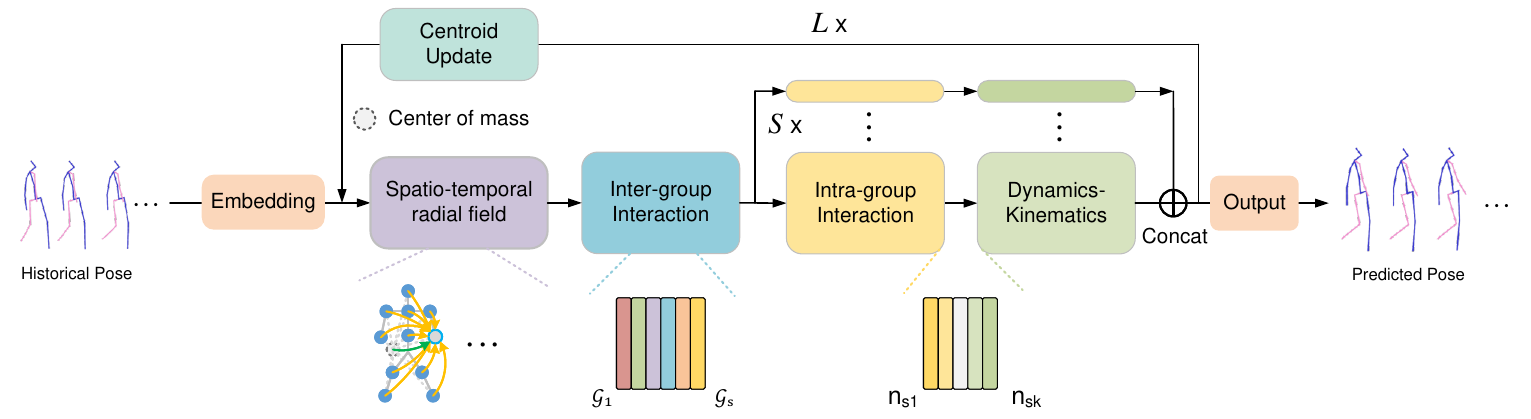}
    \caption{The proposed GGMotion architecture. We embed the physical features computed from historical poses into the network through linear layers, preserving their geometric equivariance in Euclidean space. A novel radial field is introduced to model spatial and temporal dependencies between joints. According to the characteristics of human motion, the body is independently partitioned into several groups. Inter-group and intra-group interaction modules are designed to capture joint interaction information at different scales. Joint position features are derived in parallel using the dynamics-kinematics algorithm, with the geometric centroid updated in each subsequent block. Finally, the predicted poses are obtained through a linear mapping layer.}
     \vspace{-3mm}
    \label{fig3}
\end{figure*}

\subsection{Network Architecture}
Our network adopts a sequence-to-sequence architecture based on equivariant graph neural networks. As illustrated in Fig. \ref{fig3}, motion sequences are first embedded into equivariant physical features through an embedding layer. Subsequently, spatio-temporal radial fields are employed to aggregate motion forces based on geometric edge information. To effectively model the dynamics and kinematics of the human body, the human body is partitioned into independent groups, as shown in Fig. \ref{fig1}. Inter-group and intra-group interactions are introduced to enhance joint dependencies at different scales. Joint position features are then obtained using equivariant MLPs combined with dynamics-kinematics propagation algorithms. Finally, we utilize linear layers to update the centroid and yield prediction results. In the following, we describe the design of each module in detail.

\textbf{Embedding}. We embed equivariant physical features of the \(i^{th}\) joint, including the human motion position feature \(\mathbf{X}^{0}_{i}\in\mathbb{R}^{D\times C}\) and velocity \(\mathbf{V}^{0}_{i}\in\mathbb{R}^{D\times C}\) with \(D\)-dimensional coordinates and \(C\) channels, the embedding process is:
\begin{align*}
&\mathbf{X}^{0}_{i}=\phi_{lin}(\mathbf{X}_{i}-\overline{\mathbf{X}}^{0})+\overline{\mathbf{X}}^{0}, \\
&\mathbf{V}^{0}_{i}=\phi_{lin}(\Delta^{t}\mathbf{X}_{i}),
\end{align*}
where \(\overline{\mathbf{X}}^{0}\in\mathbb{R}^{D}\) represents the centroid in time and space dimensions, \( \phi_{lin}(\cdot)\) denotes a bias-free linear layer, and \(\Delta_{t}\) indicates the difference operator to time \(t\), which corresponds to the motion velocity. 

We also utilize the sinusoidal position embedding from the Transformer to encode the number of hops (i.e., the shortest path length) between the joint \(i\) and the joint \(j\) in the skeleton topology as an edge attribute \(\tilde{\boldsymbol{d}}_{ij}\in\mathbb{R}^{C^\prime}\) to enhance the modeling of spatial relationships, where \(C^\prime\) is the hidden feature dimension used in all MLP modules.

\textbf{Spatial radial field} model the spatial dependency from joint \(i\) to joint \(j\) in Euclidean geometric space. In our method, it is implemented as:
\begin{align}
&\mathbf{e}_{ij}^{l}=\mathbf{\beta}_{i}^{l}\cdot\phi_{e}^{l}(||\mathbf{X}_{i} - \mathbf{X}_{j}||_{2,col}), \\
&\tilde{\boldsymbol{e}}^{l}_{i} = \sum_{\substack{i \in \mathcal{G}_s \\ i \neq n_{s1}}}\phi_{att}(\tilde{\boldsymbol{d}}_{n_{s1}i}) \\
&\boldsymbol{\tilde{f}}_i^{l}=\mathbf{V}_i^{l}+\tilde{\boldsymbol{e}}^{l}_{i}\cdot\sum_{j\in\mathcal{N}_i}\mathbf{e}_{ij}^{l}\cdot\phi_{lin}(\mathbf{X}_{i}^{l} - \mathbf{X}_{j}^{l}),
\end{align}
where \(\mathbf{e}_{ij}\in\mathbb{R}^{D\times C}\) is the spatial edge weight, \(\phi_{e}(\cdot)\) is implemented by MLP, \(\tilde{\boldsymbol{e}}_{i}\) is the attention weight based on each edge attribute \(\boldsymbol{d}_{ij}\). \(\phi_{att}(\cdot)\) is an attention MLP with a linear layer and a sigmoid function. \(\mathbf{\beta}_{i}^{l}\in\mathbb{R}^{D}\) is a learnable scaling factor that adaptively modulates the contribution of aggregated spatial edge features to refine spatial heterogeneity for joint \(i\). \(\boldsymbol{\tilde{f}}\) are aggregated by incorporating velocity features.

\textbf{Temporal radial fields} are utilized to capture the joint motion trajectory in Euclidean geometric space, thereby enhancing the capability for continuous frame prediction, and our algorithm is:
\begin{align}
&\mathbf{m}_{i}^{l}=\mathbf{\gamma}_{i}^{l}\cdot\phi_{m}(||\mathbf{X}_{i}^{l} - \mathbf{\overline{X}}^{l}||_{2,col}]), \\
&\boldsymbol{\overline{f}}_i^{l}=\mathbf{V}_i^{l}+\mathbf{m}_{i}^{l}\cdot\phi_{lin}(\mathbf{X}_{i}^{l} - \mathbf{\overline{X}}^{l}),
\end{align}
where \(\mathbf{m}_{i}\in\mathbb{R}^{D\times C}\) is the temporal edge weight, \(\phi_{m}(\cdot)\) is implemented by MLP. \(\mathbf{\gamma}_{i}^{l}\in\mathbb{R}^{D}\) is a learnable parameter that modulates the effect of temporal edge features, enabling the network to adaptively capture the motion trajectory features of different joints.

Ultimately, the motion force is:

\begin{align}
\boldsymbol{f}_i^{l}&=\boldsymbol{\tilde{f}}_{i}^{l} + \boldsymbol{\overline{f}}_i^{l}
\end{align}

\textbf{Inter-group interaction} module captures the motion forces acting between different body groups. We take the resultant force of the group joints as the input variable and add the residual structure, which is updated as follows:
\[\boldsymbol{f}_{i}^{l}\leftarrow\phi_{eq}([\sum_{n\in\mathcal{G}_{1}}\boldsymbol{f}_{n}^{l}, \sum_{n\in\mathcal{G}_{2}}\boldsymbol{f}_{n}^{l},\cdots, \sum_{n\in\mathcal{G}_{S}}\boldsymbol{f}_{n}^{l}])+\boldsymbol{f}_{i}^{l},\]
where \(\boldsymbol{f}_{i}^{l}\in\mathbb{R}^{D\times C}\) is the force acting on the \(i^{th}\) joint, \(\sum_{n\in\mathcal{G}_{s}}\boldsymbol{f}_{n}\) is the resultant force in the group \(\mathcal{G}_{s}\), \([\cdot,\cdot]\) is concatenation.  The output of \(\phi_{eq}(\cdot)\) is added through a skip connection to the corresponding force.

\textbf{Intra-group interaction} module deals with the dependencies between the group joints on a fine-grained scale. The operations can be described as follows:
\[\boldsymbol{f}_{i}^{l}\leftarrow\phi_{eq}^{s}([\boldsymbol{f}_{n_{s1}}^{l}, \boldsymbol{f}_{n_{s2}}^{l},\cdots,\boldsymbol{f}_{n_{sk}}^{l}])+\boldsymbol{f}_{i}^{l},\]
where \(\phi_{eq}^{s}(\cdot)\) denotes the equivariant MLP of the \(s^{th}\) group, and \(\boldsymbol{f}_{n_{sk}}^{l}\) is the force acting on the \(k^{th}\) joint in the group \(\mathcal{G}_{s}\). The skip connection is also added to the output. The module independently learns the force relationships within each group, facilitating the capture of the motion characteristics of different body parts and enabling more accurate predictions.

\textbf{Dynamics-Kinematics.} In each group, we employ an equivariant MLP to process the three fundamental physical quantities,  \(\boldsymbol{v}, \boldsymbol{r}, \boldsymbol{f} \in\mathbb{R}^{D\times C}\) as introduced in Section \ref{DK}.
Our parallel dynamics algorithm is formulated as follows:

\begin{align}
\boldsymbol{a}^{l}_{j}&=\boldsymbol{f}^{l}_{j}-\phi_{eq}([\boldsymbol{f}^{l}_{j}, \boldsymbol{r}^{l}_{ij}, \boldsymbol{v}^{l}_{ij}]), \quad i\in\mathcal{P}_{j}
\end{align}
where \(\boldsymbol{r}^{l}_{ij}=\mathbf{X}_{j}^{l} - \mathbf{X}_{i}^{l}\) and \(\boldsymbol{v}^{l}_{ij}=\mathbf{V}_{j}^{l} - \mathbf{V}_{i}^{l}\) denote the position difference and velocity difference from the parent joint \(i\) to the joint \(j\). The first term \(\boldsymbol{f}\) in the equation can be regarded as the acceleration of an isolated joint, which is subsequently refined by incorporating rigid-body motion constraints in the following terms. Compared to the hand-crafted iterative propagation in Eq. \ref{eq:aNext}, our group parallel algorithm offers a simpler solution, with superior computational efficiency. The effectiveness of this design is further validated through ablation studies reported in Section \ref{Ablation}.

The acceleration \(\boldsymbol{a}\) is further used to update velocity and position features following kinematic principles:

\begin{align}
\mathbf{V}^{l+1}_{i} &= \mathbf{V}^{l}_{i} + \phi_{lin}(\boldsymbol{a}^{l}_{i}),  \\
\mathbf{X}^{l+1}_{i} &= \mathbf{X}^{l}_{i} + \mathbf{V}^{l+1}_{i}.
\end{align}
\textbf{Centroid Update and Output}. To enhance the prediction quality, we update the centroid of the predicted positions at each layer using a linear layer without bias. The final predictions are given by:

\begin{align}
&\mathbf{\overline{X}}^{l+1}_{i} = \phi_{c}(\mathbf{X}^{l+1}_{i}),  \\
&\widehat{\mathbf{Y}} = \phi_{lin}(\mathbf{X}^{l+1}_{i} - \mathbf{\overline{X}}^{l+1}) + \mathbf{\overline{X}}^{l+1},
\end{align}
where \(\phi_{c}\) is a bias-free linear layer, followed by a mean operation to compute the \(D\)-dimensional centroid \(\mathbf{\overline{X}}\). Finally, the linear layer \(\phi_{lin}\) is applied to map the position features \(\mathbf{X}_{i}\in\mathbb{R}^{D\times C}\) to the predicted sequence \(\widehat{\mathbf{Y}}_{i}\in\mathbb{R}^{D\times T_f}\).

\subsection{Loss Function}
To supervise the kinematic constraints in human motion, in addition to the absolute position loss \(\mathcal{L}_{\text{pos}}\), we introduce joint length as an auxiliary loss \(\mathcal{L}_{\text{aux}}\). Given the past joint \(\mathbf{x}_{i}^{(t)}\), the predicted future joint \(\widehat{\mathbf{y}}_{i}^{(t)}\) and ground-truth one \(\mathbf{y}_{i}^{(t)}\), the total loss function is formulated as follows:
\begin{align*}
&\mathcal{L}_{\text{pos}}=\frac{1}{T_{\mathrm{f}}N}\sum_{t=1}^{T_f}\sum_{i=1}^{N}\|\widehat{\mathbf{y}}_{i}^{(t)}-\mathbf{y}_{i}^{(t)}\|_{2}, \\
&\mathcal{L}_{\text{aux}}=\frac{1}{T_{\mathrm{f}}(N-1)}\sum_{t=1}^{T_f}\sum_{i=1,j\in \mathcal{P}_{i}}^{N-1}\|\widehat{\mathbf{y}}_{i}^{(t)}-\mathbf{y}_{j}^{(t)}\|_{1}, \\
&\mathcal{L}=\mathcal{L}_{\text{pos}}+\mathcal{L}_{\text{aux}},
\end{align*}
where \(\mathcal{P}_{i}\) is the set of parent nodes of the joint \(i\). We use the average \(L_2\) distance for the primary loss to ensure that the future ground truth can be carefully fitted. Additionally, the auxiliary loss utilizes the average \(L_1\) distance to enhance the stability of the loss decline and accelerate convergence.

\input{tables/h36m_short}

\section{Experiment}
We evaluate on three benchmark human capture datasets to verify the effectiveness of our method. We present details of the dataset, valuation metrics, and implementation details. After that, we present the baseline and experimental results of our comparison. Exhaustive ablation experiments are presented to explain our design.

\subsection{Datasets}
The \textbf{Human3.6M} dataset~\cite{Human3.6m} includes the actions of 15 different scenes involving 7 actors. Following \cite{TrajGCN, MSRGCN}, human poses are labeled by 32 joints, and 22 joints represented by 3d coordinates are used in our experiments. With the same setup as the previous work, we downsample the pose sequence to \(25\mathrm{Hz}\), adopt \(S5\) to be the test set, \(S11\) as the validation set, and the others as the training set.

The \textbf{CMU Mocap} dataset~\cite{CMUCap} contains 8 different types of human motions, and each pose has 38 human joints. As in previous work, we discard 13 redundant joints and keep 25 joints for experiments. We similarly downsampled the sequence to \(25\mathrm{Hz}\), and maintain the partitioning of the dataset consistent with \cite{TrajGCN, MSRGCN}.

The \textbf{3DPW} dataset~\cite{3DPW} has 60 video sequences, including indoor and outdoor scenes. Following \cite{TrajGCN, PGBIGGCN}, we selected 23 joints from the original 26 joints to use and keep \(30\mathrm{Hz}\).

\subsection{Settings}
\textbf{Metrics}. The mean per joint position error (MPJPE) is widely used in evaluation metrics for 3D coordinate representation. The MPJPE calculates the average \(L_2\) distance in Euclidean space between the predicted pose and the ground truth in the same manner as \(\mathcal{L}_{\text{pos}}\), reflecting the greater pose flexibility and broader error ranges for clearer comparison.

\textbf{Implementation Details}. We set the input length \(T_h=10\) and the output length \(T_f=10/25\) on the Human3.6M dataset and CMU Mocap dataset, \(T_f=30\) on the 3DPW dataset to evaluate short-term and long-term prediction performance. The number of blocks in the network \(L\) is set to \(4\), and the number of body groups \(S\) is set to \(6\). The input sequences are scaled by a factor of \(1/1000\), and the predicted motions are rescaled to their original scale. The model is trained for \(50\) epochs with a batch size of \(64\) on a single NVIDIA GTX \(1070\) GPU. We use Adam as the optimizer during training, with an initial learning rate of \(3e^{-4}\), which decays by \(0.88\). 

\subsection{Baselines}
We compare our method with the state-of-the-art approaches on the benchmark datasets. The MPJPE is evaluated on the entire test set in a 3D coordinate space to provide a more credible and comprehensive reflection of model performance. The compared methods include Res-sup~\cite{RES-GRU}, Traj-GCN~\cite{TrajGCN}, DMGNN~\cite{DMGCN}, MSRGCN~\cite{MSRGCN}, STSGCN~\cite{STSGCN}, siMLPe~\cite{simlp}, PGBIG~\cite{PGBIGGCN}, SPGSN~\cite{SPGSNGCN}, EqMotion~\cite{Eqmotion}, DPnet~\cite{DPNet}, and KSOF~\cite{KSOFGCN}. Res-sup~\cite{RES-GRU} is a classical RNN-based method that enhances performance through residual connections. DMGNN~\cite{DMGCN} and MSRGCN~\cite{MSRGCN} abstract the body topology at multiple scales to capture richer spatial information. EqMotion~\cite{Eqmotion} employs an equivariant graph neural network with invariant interaction reasoning to learn the motion features of each node. DPnet~\cite{DPNet} proposes a dynamic pattern-based collaborative modeling approach to capture joint dynamic information. In addition to GCN, STSGCN~\cite{STSGCN} and KSOF~\cite{KSOFGCN} incorporate the Temporal Convolutional Network (TCN) to capture temporal dependencies. For a fair comparison, we either use their publicly available pre-trained models or re-train them with the default hyperparameters.

\input{tables/h36m_long}

\begin{figure*}
    \centering
    \includegraphics[width=1\linewidth]{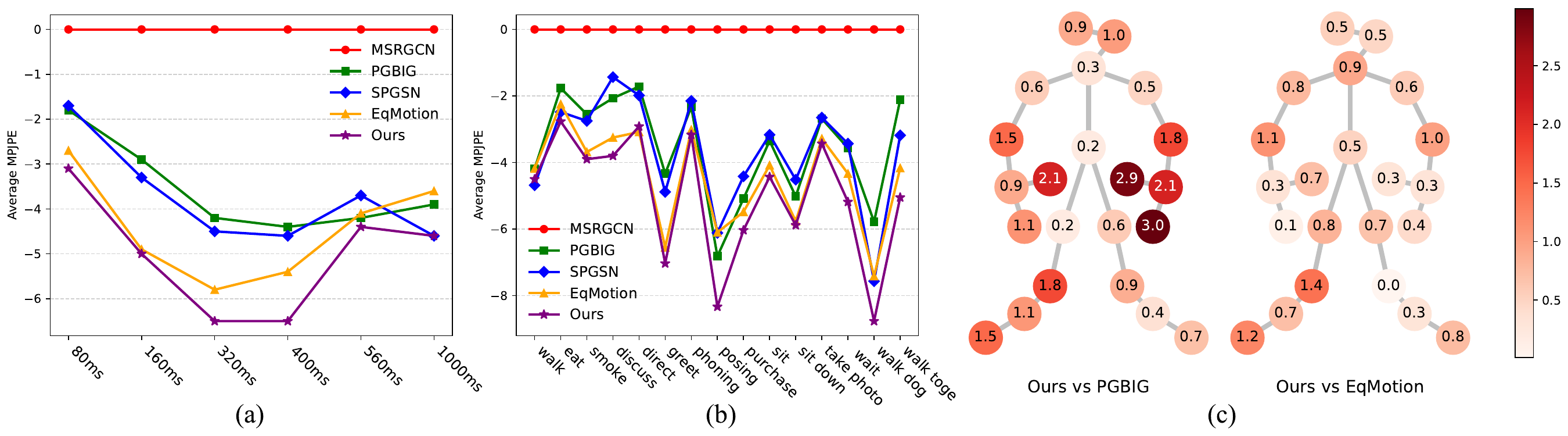}
    \vspace{-6mm}
    \caption{Intuitive comparison on Human3.6M. (a) Average Performance gain at different timestamps. (b) Average Performance gains on different actions. (c) Average performance gains on different joints.}
     \vspace{-3mm}
    \label{fig4}
\end{figure*}

\subsection{Results}
We provide the qualitative and quantitative results of our approach in both short-term (i.e., \(\leq 500\,\mathrm{ms}\)) and long-term (i.e., \(> 500\,\mathrm{ms}\)) prediction.

\textbf{Results on Human3.6M.} We present qualitative comparisons of short-term motion predictions on Human3.6M with several existing methods in Table \ref{tab:h36m_short}. In short-term predictions, our method achieves leading performance in most cases and outperforms baseline methods by a large margin on average. In long-term predictions, our method encounters challenges arising from greater temporal uncertainty and high motion frequency. Although not utilizing temporal modeling methods such as TCN~\cite{STSGCN, KSOFGCN} or DCT~\cite{TrajGCN, MSRGCN, PGBIGGCN, SPGSNGCN}, which are known to benefit sequence prediction, our method still delivers competitive prediction results, as demonstrated in Table \ref{tab:h36m_long}.

\input{tables/cmu}

In addition, we analyze the statistical results and provide an intuitive comparison with other methods in Fig. \ref{fig4}. In (a), we compare the average MPJPE of baseline methods across different timestamps. We take MSRGCN as the baseline and compute the relative prediction errors of other methods. Our prediction curve lies at the bottom among all curves and achieves a significant advantage in \(320ms\) and \(400ms\). Our method substantially surpasses EqMotion~\cite{Eqmotion} in \(1000ms\), which only aggregates information from the spatial neighboring edges, highlighting the contribution of the proposed spatio-temporal radial field to improving the accuracy of long-term prediction. In (b), we plot the average prediction error gains relative to MSRGCN across all timestamps for each action category. As can be seen, our method outperforms other methods with notably larger improvements on actions like "posing" and "walking together". On the other hand,  we analyze the average MPJPE for each joint and present the performance gains relative to PGBIG and EqMotion in (c). The advantage of our method is more significant for joints near the extremities, highlighting the effectiveness of the proposed dynamics-kinematics algorithm.

\input{tables/3dpw}

\textbf{Results on CMU MoCap and 3DPW.} The same comparisons on CMU MoCap and 3DPW are presented in Table \ref{tab:CMU} and Table \ref{tab:3DPW} respectively. Compared to Human3.6M, our method performs better on CMU Mocap and 3DPW, with additional MPJPE reductions of \(0.9/3.5mm\) at \(400ms\). This is likely attributed to differences in body topology and joint sampling accuracy, where a more realistic joint distribution results in improved prediction accuracy. Relative to the dynamic pattern-based collaborative modeling in DPnet~\cite{DPNet}, our refined grouping strategy and dynamic-kinematic modeling provide notable improvements.
Consistent improvements across all benchmarks demonstrate the generalizability and reliability of our human topology modeling strategy.

\begin{figure*}
    \centering
    \includegraphics[width=1\linewidth]{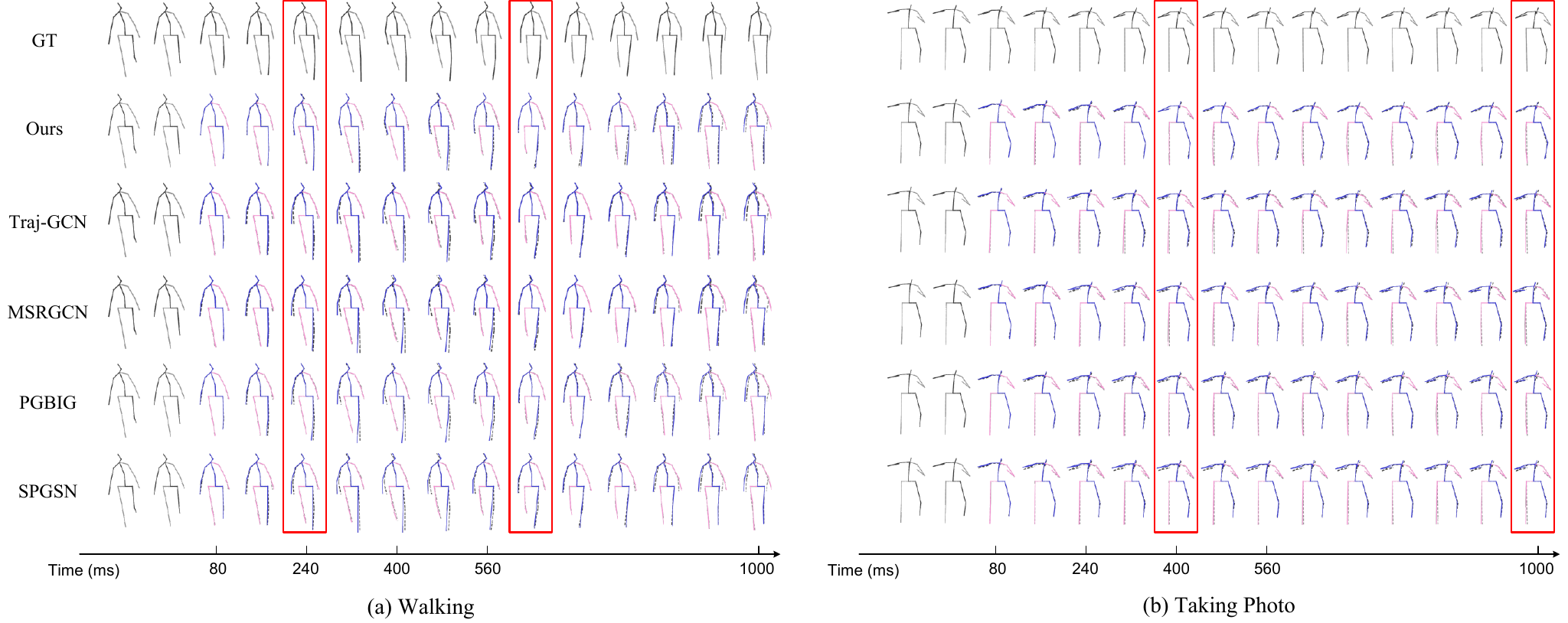}
    \vspace{-6mm}
    \caption{Visualization of predicted poses of different methods on Human3.6M. The first row presents the ground truth in gray, while the second row displays our predicted results in color. The predicted poses where our method achieves more accurate predictions are highlighted with red boxes.}
     \vspace{-3mm}
    \label{fig5}
\end{figure*}

\textbf{Qualitative results.}
We present a qualitative example of predicted poses from different methods in Fig. \ref{fig5}. It can be seen that our method produces more accurate short-term predictions that closely match the ground truth, especially near the body extremities. In long-term predictions, while the performance gap between our approach and the baselines is marginal, our method generates more credible and stable human motion poses. From the perspective of motion scenarios, our method achieves superior results in both the large-amplitude action "walking" and the small-amplitude action "taking photo".

\textbf{Efficiency results.}
We report the practical applicability of different approaches on Human3.6M in Table \ref{tab:model_size}. To ensure a fair comparison, all methods are tested with the same batch size, and the model size is reported for long-term prediction tasks. Our approach achieves both the lowest MPJPE and the smallest model size, which is attributed to the maintained equivariance in the predicted results. Meanwhile, the running time remains in an acceptable range.

\input{tables/mode_size}
\input{tables/ablation_module}
\input{tables/ablation_block}
\input{tables/ablation_group}
\input{tables/ablation_loss}

\section{Ablation Study}
\label{Ablation}
To further validate the effect of our approach, we perform ablation experiments on Human3.6M.

\textbf{Effect of key modules.}
We conduct ablation experiments on key modules of our network, as shown in Table \ref{tab:ablation_module}. (1) Removing the "centroid update" alone led to increased prediction error at all timestamps, confirming the necessity of centroid updates in each block. (2) To validate the effectiveness of the proposed spatio-temporal radial fields, we conducted ablation studies by individually retaining the spatial component (“S. field”) and the temporal component (“T. field”) to observe their impact on prediction errors. In short-term predictions, the “T. field” achieved lower prediction errors than the “T. field”, highlighting the crucial role of temporal fields in capturing joint motion trajectories. We further removed the learnable scaling factors from the spatio-temporal radial fields, which also led to a slight increase in prediction errors. Then we removed the entire spatio-temporal radial field and replaced the motion forces with velocities in the subsequent processes. As a result, prediction errors increased more significantly, demonstrating the essential role of the spatio-temporal fields in capturing joint dependencies. (3) We conducted ablations by removing the inter-group and intra-group interaction modules separately. The inter-group interaction, modeling correlations among body parts, had a greater impact on performance than the intra-group interaction. (4) We manually implement the iterative dynamics-kinematics algorithm in Eq. \ref{eq:aNext}, and observe that its short-term prediction performance is inferior to the direct output. This finding further verifies the simplicity and efficiency of our parallel prediction approach. (5) Replacing our proposed equivariant MLP with the MLP without an attention mechanism in \cite{GMN}  led to a significant increase in the prediction error, confirming the effectiveness of our equivariant MLP in modeling continuous motion sequences.

\textbf{Number of blocks.}
The impact of the number of network blocks \(L\) is evaluated through ablation experiments, as shown in Table \ref{tab:ablation_block}. With four blocks, the network achieves the best performance on short-term prediction tasks, while preserving a compact parameter size of \(0.46M\). The experimental results show that the network performance improves as the number of blocks increases, achieving the best results when \(L=4\). 

\textbf{Effect of the group strategy.}
To validate the effectiveness of the grouping strategy in the dynamics-kinematics propagation algorithm, we divide the human body into three hierarchical schemes:
(1) “1”: treating the entire body as a single group, which is equivalent to disabling the grouping strategy;
(2) “2”: following the grouping strategy from \cite{SPGSNGCN, Skeleton-Aware} to divide the human body into upper and lower parts;
(3) “5”: dividing the body into five groups by merging the head and spine into a single group based on our method;
(4) “6”: following our proposed six-group configuration as shown in Fig. \ref{fig1}.
The experimental results are reported in Table \ref{tab:ablation_group}. It can be observed that the six-group configuration achieves the best performance, which is consistent with the natural motion characteristics of different body parts.

\textbf{Effect of Auxiliary Loss.}
In Table \ref{tab:ablation_loss}, we evaluate the contribution of different loss terms during training. As shown, incorporating the auxiliary loss \(\mathcal{L}_{\text{aux}}\) leads to the best prediction performance at all timestamps. These results demonstrate the crucial role of incorporating the human skeletal length loss term in supervising the training process.

\section{Conclusion}
In this work, we present a novel group graph dynamics-kinematics neural network (GGMotion) for human motion prediction. A novel radial field is introduced to adaptively capture the spatio-temporal dependencies of each node through neighboring and centroid edges, while preserving the equivariant geometric transformations of human motion. The group strategy is employed to construct a comprehensive human motion topology compatible with the motion characteristics of different body parts. To further enhance motion plausibility, we propose a parallel dynamics-kinematics algorithm in combination with the auxiliary loss. Our model exhibits geometric equivariance and achieves competitive performance with fewer parameters. Extensive experiments on three standard benchmark datasets confirm the effectiveness of our approach.
\bibliographystyle{IEEEtran}
\bibliography{ref}
\vfill

\end{document}

%% file: tables/h36m_short.tex
\begin{table*}[t]
\caption{\small Comparisons of short-term prediction on Human3.6M. Results at 80ms, 160ms, 320ms and 400ms in the future are shown. All samples in the test set are tested for each action. The best results are highlighted in \textbf{bold}, and the second best result is \underline{underline}.}
\vspace{-3mm}
\renewcommand\arraystretch{0.95}
\resizebox{1.0\textwidth}{!}{
\scriptsize
\begin{tabular}{c|cccc|cccc|cccc|cccc} \hline
Motion & \multicolumn{4}{c|}{Walking} & \multicolumn{4}{c|}{Eating} & \multicolumn{4}{c|}{Smoking} & \multicolumn{4}{c}{Discussion} \\ \hline
millisecond & 80ms & 160ms & 320ms & 400ms & 80ms & 160ms & 320ms & 400ms & 80ms & 160ms & 320ms & 400ms & 80ms & 160ms & 320ms & 400ms \\ \hline
Res-sup \cite{RES-GRU} & 29.4 & 50.8 & 76.0 & 81.5 & 16.8 & 30.6 & 56.9 & 68.7 & 23.0 & 42.6 & 70.1 & 82.7 & 32.9 & 61.2 & 90.9 & 96.2 \\
Traj-GCN \cite{TrajGCN} & 12.3 & 23.0 & 39.8 & 46.1 & 8.4  & 16.9          & 33.2          & 40.7          &  7.9           &  16.2          & 31.9          & 38.9          & 12.5          & 27.4          & 58.5          & 71.7          \\
STSGCN \cite{STSGCN} &16.3 &24.6 &40.1 &45.9 &14.3 &22.1 &37.9 &45.0 &13.1 &20.2 &37.7 &44.7 &14.3 &24.3 &52.6 &68.5 \\
DMGNN \cite{DMGCN}            & 17.3          & 30.7          & 54.6          & 65.2          & 11.0          & 21.4          & 36.2          & 43.9          & 9.0           & 17.6          & 32.1          & 40.3          & 17.3          & 34.8          & 61.0          & 69.8          \\
MSRGCN \cite{MSRGCN}            &  12.2          &  22.7          &  38.6          &  45.2          &  8.4           & 17.1          &  33.0          &  40.4          & 8.0           & 16.3          &  31.3          &  38.2          &  12.0          &  26.8          &  57.1          &  69.7          \\
siMLPe \cite{simlp}  &10.3 &21.0 &38.9 &46.2   &\textbf{6.2} &14.2 &\underline{29.5} &\underline{37.1}   &6.8 &15.0 &31.4 &38.7  &9.6 &23.1 &51.8 &65.5\\
PGBIG \cite{PGBIGGCN}            &   10.2          &   19.8          &    {34.5}          &  40.3          &{7.0}  & 15.1         & 30.6          &   {38.1}          &   {6.6}           &   {14.1}          &   {28.2}          &   {34.7}          &   {10.0}          &   {23.8}          &   {53.6}          &   {66.7}          \\ 
SPGSN \cite{SPGSNGCN}             &   {10.1}          &   {19.4}          &   {34.8}          &   {41.5}          &   {7.1}           &   {14.9}          &   {30.5}          &   {37.9}          &   {6.7}           &   {13.8}          &   {28.0}          &   {34.6}          &   {10.4}          &   {23.8}          &   {53.6}          &   {67.1}          \\
EqMotion \cite{Eqmotion}  
&\underline{9.3} &\textbf{17.6} &\textbf{32.9} &\underline{39.9}
&\underline{6.5} &\underline{13.9} &30.0 &37.8
&\underline{6.0} &\textbf{12.6} &26.6 &33.6
&\underline{9.1} &\underline{22.0} &52.1 &65.8
 \\

KSOF \cite{KSOFGCN}    
&9.4 &19.1 &35.3 &\textbf{38.2} 
&6.6 &14.7 &30.6 &37.8 
&6.2 &13.7 &\textbf{25.3} &\textbf{31.2} 
&9.3 &23.0 &\textbf{48.1} &\textbf{60.2} \\
GGMotion (Ours) 
&\textbf{9.1}  &\underline{18.0} &\underline{33.4} &40.3
&\textbf{6.2}  &\textbf{13.8} &\textbf{29.4} &\textbf{36.8}
&\textbf{5.8}  &\underline{12.8} &\underline{26.2} &\underline{33.2} 
&\textbf{8.7}  &\textbf{21.8} &\underline{51.3} &\underline{64.5} \\
\hline
Motion & \multicolumn{4}{c|}{Directions} & \multicolumn{4}{c|}{Greeting} & \multicolumn{4}{c|}{Phoning} & \multicolumn{4}{c}{Posing} \\ \hline
millisecond & 80ms & 160ms & 320ms & 400ms & 80ms & 160ms & 320ms & 400ms & 80ms & 160ms & 320ms & 400ms & 80ms & 160ms & 320ms & 400ms \\ \hline
Res-sup \cite{RES-GRU}         & 35.4          & 57.3          & 76.3          & 87.7          & 34.5          & 63.4          & 124.6         & 142.5         & 38.0          & 69.3          & 115.0         & 126.7         & 36.1          & 69.1          & 130.5         & 157.1         \\
Traj-GCN \cite{TrajGCN}               & 9.0           & 19.9          & 43.4          &  53.7          & 18.7          & 38.7          & 77.7          & 93.4          & 10.2          & 21.0          & 42.5          & 52.3          & 13.7          & 29.9          &  66.6          &  84.1          \\
STSGCN \cite{STSGCN} &14.2 &24.3 &44.3 &53.2 &15.0 &30.7 &67.1 &87.3 &14.88 &21.40 &46.55 &52.0 &15.0 &25.7 &67.0 &85.0 \\
DMGNN \cite{DMGCN}             & 13.1          & 24.6          & 64.7          & 81.9          & 23.3          & 50.3          & 107.3         & 132.1         & 12.5          & 25.8          & 48.1          & 58.3          & 15.3          & 29.3          & 71.5          & 96.7  \\
MSRGCN \cite{MSRGCN}               &  8.6           &  19.7          &  43.3          & 53.8          &  16.5          &  37.0          &  77.3          &  93.4          &  10.1          &  20.7          &  41.5          &  51.3          &  12.8          &  29.4          & 67.0          & 85.0          \\
siMLPe \cite{simlp}   &6.8 &17.8 &44.3 &56.2  &\underline{12.7} &\textbf{29.3} &\textbf{64.8} &\textbf{79.9}  &8.3 &18.5 &40.4 &50.8  &\underline{9.1} &23.5 &57.9 &74.4\\
PGBIG \cite{PGBIGGCN}               &   {7.2}  &   {17.6} &   {40.9} &  {51.5} &   {15.2} &   {34.1} &   {71.6} &   {87.1} &  {8.3}  &   {18.3} &   {38.7} &  {48.4} &   {10.7} &   {25.7} &   {60.0} &   {76.6} \\ 
SPGSN \cite{SPGSNGCN}   & 7.4          & {17.2}  & \underline{39.8}  & {50.3}  & {14.6}  & {32.6}          &  {70.6}         &  {86.4}         & 8.7  & {18.3} & {38.7}  & 48.5   & {10.7} & {25.3} & {59.9} & {76.5}               \\ 
EqMotion \cite{Eqmotion}
&\underline{6.5} &\textbf{15.6} &\textbf{38.5} &\underline{49.5}
&12.8 &29.9 &68.8 &85.8
&\underline{7.7} &\textbf{16.9} &\underline{37.7} &48.0
&9.4 &23.5 &59.4 &77.3    \\
KSOF \cite{KSOFGCN}  
&6.6 &17.0 &40.4 &49.8   
&13.4 &31.4 &\underline{67.7} &85.9    
&7.8 &18.0 &37.9 &\underline{47.7}    
&9.2 &\textbf{23.1} &\underline{57.4} &\underline{74.1}   \\
GGMotion (Ours) 
&\textbf{6.1} &\underline{15.7} &\textbf{38.5} &\textbf{49.3}
&\textbf{12.4} &\underline{29.8} &\underline{67.7} &\underline{84.0} 
&\textbf{7.4} &\underline{17.0} &\textbf{37.3} &\textbf{47.1}
&\textbf{9.0} &\underline{23.2} &\textbf{57.2} &\textbf{74.0}
\\ \hline
Motion           & \multicolumn{4}{c|}{Purchases}                                 & \multicolumn{4}{c|}{Sitting}                                   & \multicolumn{4}{c|}{Sittingdown}                               & \multicolumn{4}{c}{Takingphoto}                               \\ \hline
millisecond        & 80ms          & 160ms         & 320ms         & 400ms         & 80ms          & 160ms         & 320ms         & 400ms         & 80ms          & 160ms          & 320ms          & 400ms          & 80ms           & 160ms          & 320ms          & 400ms          \\ \hline
Res-sup \cite{RES-GRU}         & 36.3          & 60.3          & 86.5          & 95.9          & 42.6          & 81.4          & 134.7         & 151.8         & 47.3          & 86.0          & 145.8         & 168.9         & 26.1          & 47.6          & 81.4          & 94.7          \\
Traj-GCN \cite{TrajGCN}              & 15.6          & 32.8          & 65.7          &  79.3          & 10.6          &  21.9          & 46.3          & 57.9          & 16.1          &  31.1          &  61.5          &  75.5          & 9.9           &  20.9          & 45.0          & 56.6          \\
STSGCN \cite{STSGCN} &15.3 &\underline{26.3} &63.5 &74.3 &15.2 &23.0 &46.8 &58.3 &16.7 &28.0 &56.2 &72.0 &16.6 &24.8 &46.0 &61.8 \\
DMGNN \cite{DMGCN}             & 21.4          & 38.7          & 75.7          & 92.7          & 11.9          & 25.1          & 44.6          &   \textbf{50.2}         & 15.0          & 32.9          & 77.1          & 93.0          & 13.6          & 29.0          & 46.0          & 58.8  \\
MSRGCN \cite{MSRGCN}       &  14.8          &  32.4          & 66.1          & 79.6          &  10.5          & 22.0          &  46.3          & 57.8          &  16.1          & 31.6          & 62.5          & 76.8          &  9.9           & 21.0          &  44.6          &  56.3         \\
siMLPe \cite{simlp}  &12.0 &28.4 &61.1 &75.3   &9.0 &20.1 &45.0 &56.8 &13.9 &30.1 &58.7 &71.5   &8.1 &18.2 &41.0 &\underline{52.1}    \\
PGBIG \cite{PGBIGGCN}               &   {12.5} &   {28.7} &   {60.1} &   {73.3}    &   {8.8}  &   {19.2} &   {42.4} &  53.8 &   {13.9} &   {27.9} &   {57.4} &   {71.5} &   {8.4}  &   {18.9} &   {42.0} &   {53.3} \\ 
SPGSN \cite{SPGSNGCN}               & 12.8 &{28.6} &{61.0}  & {74.4}     & 9.3  & 19.4 & {42.3} &  {53.6}    & 14.2 &{27.7}  &{56.8}  & {70.7}       & 8.8  & {18.9}  &{41.5} & {52.7}              \\ 
EqMotion \cite{Eqmotion}   
&\underline{11.4} &26.4 &\underline{58.8} &\underline{72.9}
&\underline{8.1} &\textbf{17.9} &\textbf{40.9} &52.4
&\underline{13.0} &\underline{26.2} &\underline{55.6} &\underline{70.0}
&\underline{8.0} &\textbf{17.6} &\underline{40.8} &52.5
\\
KSOF \cite{KSOFGCN}   
&11.7 &28.1 &61.5 &75.1   
&8.4 &19.0 &42.2 &\underline{51.4}  
&13.8 &28.4 &58.1 &72.3  
&9.2 &18.7 &42.0 &55.3   \\
GGMotion (Ours) 
&\textbf{10.7} &\textbf{25.9} &\textbf{57.8} &\textbf{71.4} 
&\textbf{8.0}  &\underline{18.0}  &\underline{41.1}  &53.1
&\textbf{12.8}  &\textbf{26.0}  &\textbf{55.1}  &\textbf{69.2}
&\textbf{7.7}  &\underline{17.7}  &\textbf{40.3} &\textbf{51.4}
\\\hline
Motion           & \multicolumn{4}{c|}{Waiting}                                   & \multicolumn{4}{c|}{Walking Dog}                                & \multicolumn{4}{c|}{Walking Together}                           & \multicolumn{4}{c}{Average}                                       \\ \hline
millisecond       & 80ms          & 160ms         & 320ms         & 400ms         & 80ms          & 160ms         & 320ms         & 400ms         & 80ms          & 160ms          & 320ms          & 400ms          & 80ms           & 160ms          & 320ms          & 400ms          \\ \hline
Res-sup \cite{RES-GRU}         & 30.6          & 57.8          & 106.2         & 121.5         & 64.2          & 102.1         & 141.1         & 164.4         & 26.8          & 50.1          & 80.2          & 92.2          & 34.7          & 62.0          & 101.1         & 115.5         \\
Traj-GCN \cite{TrajGCN}               & 11.4          & 24.0          & 50.1          & 61.5          & 23.4          & 46.2          & 83.5          & 96.0          &  10.5          & 21.0          & 38.5          & 45.2          & 12.7          & 26.1          & 52.3          & 63.5          \\
STSGCN \cite{STSGCN} &16.3 &27.3 &48.1 &59.8 &16.5 &37.6 &\underline{70.6} &86.3 &11.4 &22.4 &39.9 &47.5 &15.3 &25.5 &50.6 &60.6 \\
DMGNN \cite{DMGCN}             & 12.2          & 24.2          & 59.6          & 77.5          & 47.1          & 93.3          & 160.1         & 171.2         & 14.3          & 26.7          & 50.1          & 63.2          & 17.0          & 33.6          & 65.9          & 79.7          \\
MSRGCN \cite{MSRGCN}               &  10.7          &  23.1          &  48.3          & 59.2          &  20.7          &  42.9          &  80.4          &  93.3          & 10.6          &  20.9          &  37.4         &  43.9          &  12.1          &  25.6          &  51.6          &  62.9         \\
siMLPe \cite{simlp}  &8.0 &18.4 &43.2 &55.0 &18.2 &38.4 &72.5 &85.8 &8.7 &19.0 &37.6 &45.4 &9.9 &22.3 &47.9 &59.4    \\
PGBIG \cite{PGBIGGCN}      &   8.9  &   20.1 & 43.6 &  54.3 &  18.8 &  39.3 & 73.7 &   86.4 &   8.7  &  18.6 &   34.4 &   41.0 &   10.3  &  22.7  &   47.4  &   58.5 \\ 
SPGSN \cite{SPGSNGCN}      & 9.2  & 19.8  & 43.1  & 54.1  & 17.8   & 37.2   & 71.7  & 84.9      & 8.9   &18.2  & 33.8  & 40.9     & 10.4  &22.3  &47.1  & 58.3         \\ 
EqMotion \cite{Eqmotion}   
&\underline{7.9} &\underline{17.8} &\underline{41.2} &\underline{52.9}
&16.6 &\underline{35.9} &71.8 &85.3
&8.2 &\textbf{16.7} &\underline{32.0} &\underline{39.0}
&\underline{9.4} &\underline{20.7} &\underline{45.8} &57.5
\\
KSOF \cite{KSOFGCN}  
&8.2 &19.3 &43.0 &53.6 
&\textbf{15.9}  &36.9 &72.2 &\underline{84.7}  
&\textbf{7.8} &18.0 &34.0 &40.5   
&9.5 &21.3 &46.1 &\underline{57.4}   \\
GGMotion (Ours) 
&\textbf{7.6}  &\textbf{17.6}  &\textbf{40.5}  &\textbf{51.7}
&\underline{16.2}  &\textbf{35.2}  &\textbf{69.0}  &\textbf{82.0}
&\underline{8.0}  &\underline{17.0}  &\textbf{31.6}  &\textbf{38.7} 
&\textbf{9.0}  &\textbf{20.6}  &\textbf{45.1}  &\textbf{56.4}
\\\hline
\end{tabular}}
\vspace{-3mm}
\label{tab:h36m_short}
\end{table*}

%% file: tables/h36m_long.tex
\begin{table*}[ht]
\caption{\small Comparisons of long-term prediction on Human3.6M. Results at 560ms and 1000ms in the future are shown. The best results are highlighted in \textbf{bold}, and the second best result is \underline{underlined}.}
\vspace{-3mm}
\renewcommand\arraystretch{0.95}
\resizebox{\textwidth}{!}{
\begin{tabular}{c|cc|cc|cc|cc|cc|cc|cc|cc} \hline
Motion & \multicolumn{2}{c|}{Walking}    & \multicolumn{2}{c|}{Eating}     & \multicolumn{2}{c|}{Smoking}     & \multicolumn{2}{c|}{Discussion}  & \multicolumn{2}{c|}{Directions} & \multicolumn{2}{c|}{Greeting}    & \multicolumn{2}{c|}{Phoning}         & \multicolumn{2}{c}{Posing}      \\ \hline
millisecond        & 560ms         & 1000ms         & 560ms         & 1000ms         & 560ms          & 1000ms         & 560ms          & 1000ms         & 560ms         & 1000ms         & 560ms          & 1000ms         & 560ms            & 1000ms           & 560ms          & 1000ms         \\ \hline
Res-sup \cite{RES-GRU}          & 81.7          & 100.7          & 79.9          & 100.2          & 94.8           & 137.4          & 121.3          & 161.7          & 110.1         & 152.5          & 156.3          & 184.3          & 143.9            & 186.8            & 165.7          & 236.8          \\
Traj-GCN \cite{TrajGCN}               & 54.1          & 59.8           & 53.4          & 77.8           & 50.7           & 72.6           & 91.6           & 121.5          &   {71.0}          & 101.8          &   {115.4}          & 148.8          & 69.2             & 103.1            &   {114.5}          &   {173.0}          \\
STSGCN \cite{STSGCN} &53.8 &63.5 &57.4 &81.5 &51.5 &72.8 &92.3 &120.1 &74.9 &103.4 &115.6 &145.2 &72.2 &107.0 &118.7 &171.2 \\
DMGNN \cite{DMGCN} &73.4 &95.8 &58.1 &86.7 &50.9 &72.2 &81.9 &138.3 &110.1 &115.8 &152.5 &157.7 &78.9 &\textbf{98.6} &163.9 &310.1 \\
MSRGCN \cite{MSRGCN}               &   {52.7}          &   {63.0}           &   {52.5}          &   {77.1}           &   {49.5}           &   {71.6}           &   {88.6}           &   {117.6} & 71.2          &   {100.6}          & 116.3          &   {147.2}          &   {68.3}             & 104.4            & 116.3          & 174.3          \\
siMLPe \cite{simlp}  &56.1 &64.6 &50.6 &76.4   &51.0 &72.7 &87.9 &119.6    &73.8 &107.2 &\textbf{102.8} &\textbf{137.7}  &69.2 &106.1 &\textbf{104.8} &171.7          \\
PGBIG  \cite{PGBIGGCN}   &   {48.1} &   {56.4}  &   {51.1} &   {76.0}  &   {46.5}  & 69.5  &   {87.1}  &   {118.2}          & {69.3} &   {100.4} &   {110.2} &   {143.5} &   \textbf{65.9}    &   {102.7}   &   \underline{106.1} &   \textbf{164.8} \\ 
SPGSN \cite{SPGSNGCN} &\underline{46.9} &\underline{53.6} & \textbf{49.8}& \textbf{73.4}& 46.7 &\underline{68.6} & 89.7& 118.6& {70.1}& 100.5& 111.0& {143.2}& 66.7& 102.5& 110.3& 165.4 \\
EqMotion \cite{Eqmotion} 
&50.4 &59.2 &50.9 &75.9
&\underline{46.1} &\textbf{67.9} &\underline{86.3} &117.0
&\underline{68.8} &\underline{99.8} &\underline{108.6} &\underline{142.5}
&66.2 &\underline{101.7} &109.4 &169.2
\\
KSOF \cite{KSOFGCN}   
&\textbf{42.3} &\textbf{49.2} &52.4 &75.7
&48.4 &70.2 &\textbf{78.4} &\textbf{111.3} 
&\textbf{68.2} &\textbf{99.5} &113.2 &145.3 
&67.6 &102.9 &109.4 &173.6\\
GGMotion (Ours)
&50.0 &56.6 &\underline{50.4} &\underline{75.3}  
&\textbf{45.6} &\textbf{67.9} &86.8  &\underline{115.9}
&69.6  &100.5  &109.1  &\underline{142.5}
&\underline{66.0}  &102.5  &106.4  &\underline{165.0}
\\\hline
Motion          & \multicolumn{2}{c|}{Purchases}  & \multicolumn{2}{c|}{Sitting}    & \multicolumn{2}{c|}{Sitting Down} & \multicolumn{2}{c|}{Taking Photo} & \multicolumn{2}{c|}{Waiting}    & \multicolumn{2}{c|}{Walking Dog}  & \multicolumn{2}{c|}{Walking Together} & \multicolumn{2}{c}{Average}     \\ \hline
millisecond        & 560ms         & 1000ms         & 560ms         & 1000ms         & 560ms          & 1000ms         & 560ms          & 1000ms         & 560ms         & 1000ms         & 560ms          & 1000ms         & 560ms            & 1000ms           & 560ms          & 1000ms         \\ \hline
Res-sup \cite{RES-GRU}          & 119.4         & 176.9          & 166.2         & 185.2          & 197.1          & 223.6          & 107.0          & 162.4          & 126.7         & 153.2          & 173.6          & 202.3          & 94.5            & 110.5            & 129.2        & 165.0          \\
Traj-GCN \cite{TrajGCN}               & 102.0         & 143.5          & 78.3          & 119.7          &   {100.0}          &   {150.2}          &   {77.4}           &   {119.8}          & 79.4          & 108.1          & 111.9          & 148.9          & 55.0             &   {65.6}             & 81.6           & 114.3          \\
STSGCN \cite{STSGCN}  &103.9 &140.3 &82.4 &121.3 &105.1 &153.6 &82.9 &122.6 &80.8 &109.6 &111.7 &145.3 &55.0 &66.5 &83.9 &114.9 \\
DMGNN \cite{DMGCN} &118.6 &153.8 &\textbf{60.1} &\textbf{104.9} &122.1 &168.8 &91.6 &120.7 &106.0 &136.7 &194.0 &182.3 &83.4 &115.9 &103.0 &137.2 \\
MSRGCN \cite{MSRGCN}               &   {101.6}         &   {139.2}          & 78.2          & 120.0          & 102.8          & 155.5          & 77.9           & 121.9          &   {76.3}          &   {106.3}          &   {111.9}          &   {148.2}          &   {52.9}             & 65.9             &   {81.1}           &   {114.2}          \\
siMLPe \cite{simlp}  &97.0 &137.2 &77.8 &117.1 &\textbf{96.0} &\textbf{142.2} &\textbf{72.8} &\underline{115.2} &74.4 &107.6 &106.5 &145.7 &57.8 &69.4 &78.6 &112.7         \\
PGBIG \cite{PGBIGGCN}     &  \textbf{95.3} &   \underline{133.3} &   74.4 &   116.1 &\underline{96.7}  &  \underline{147.8} &   \underline{74.3}  &   {118.6} &   \textbf{72.2} &   \textbf{103.4} &   {104.7} &   {139.8} &   {51.9}    &   {64.3}    &    \underline{76.9}  &   {110.3} \\ 
SPGSN \cite{SPGSNGCN} & 96.5& 133.9& 75.0& 116.2& 98.9& 149.9& 75.6& 118.2 & 73.5 & 103.6 & \underline{102.4} & \underline{138.0} & \underline{49.8} & 60.9 & 77.4 & \textbf{109.6}\\
EqMotion \cite{Eqmotion} 
&\textbf{95.3} &136.0 &74.5 &116.5
&\underline{96.7} &149.2 &74.6 &118.4
&72.8 &105.3 &103.6 &139.7
&50.6 &\underline{60.1} &77.0 &110.6
\\
KSOF \cite{KSOFGCN}  
&97.5 &\textbf{133.2} &82.7 &116.2 
&99.7 &151.9 &77.4 &\textbf{113.2}
&74.1 &105.1 &\textbf{100.8} &\textbf{134.2}
&53.3 &63.8 &77.7 &\underline{109.7}   \\
GGMotion (Ours) 
&\underline{95.9}  &135.8  &\underline{73.5}  &\underline{114.5}
&97.3  &149.6  &75.3  &118.6
&\underline{72.5}  &\textbf{102.9}  &103.4  &139.0
&\textbf{48.5}  &\textbf{57.5}  &\textbf{76.7}  &\textbf{109.6}
\\\hline

\end{tabular} 
}
\vspace{-3mm}
\label{tab:h36m_long}
\end{table*}

%% file: tables/cmu.tex
\begin{table}[t]
      \small 
      \centering
      \caption{\small Comparisons of average prediction errors on CMU.}
      \vspace{-3mm}
       \renewcommand\arraystretch{0.93}
      \resizebox{\columnwidth}{!}{
      \begin{tabular}{c|ccccc}
      \hline
      millisecond & 80 & 160 & 320  & 400 & 1000 \\
      \hline
      Res-sup.~\cite{RES-GRU} &24.0 &43.0 &74.5 &87.2  &136.3 \\
      Traj-GCN~\cite{TrajGCN} &9.3 &17.1 &33.0 &40.9 &86.2 \\
      DMGNN~\cite{DMGCN} &13.6 &24.1 &47.0 &58.8  &112.6 \\
      MSR-GCN~\cite{MSRGCN} &8.1 &15.2 &30.6 &38.6 &83.0 \\
      STSGCN~\cite{STSGCN}  &10.8 &18.2 &31.2 &41.2 &81.7 \\
      PGBIG \cite{PGBIGGCN} &\underline{7.6} &\underline{14.3} &29.0 &\underline{36.6} &80.1 \\
      SPGSN \cite{SPGSNGCN}  &8.3 &14.8 &\underline{28.6} &37.0 &\textbf{77.8} \\
      DPnet \cite{DPNet} &8.4 &14.5 &30.6 &39.7&91.3 \\
    GGMotion (Ours) &\textbf{7.1}  &\textbf{13.2}  &\textbf{27.0} &\textbf{34.7} &\underline{78.6} \\\hline
      \end{tabular}}
      \vspace{-3mm}
      \label{tab:CMU}
\end{table}

%% file: tables/3dpw.tex
\begin{table}[t]
      \small 
      \centering
      \caption{\small Comparisons of average prediction errors on 3DPW.}
      \vspace{-3mm}
      \renewcommand\arraystretch{1.00}
      \resizebox{\columnwidth}{!}{
      \begin{tabular}{c|ccccc}
      \hline
      millisecond & 200 & 400  & 600 & 800  & 1000 \\
      \hline
      Res-sup.~\cite{RES-GRU} &113.9 &173.1 &191.9 &201.1 &210.7 \\
      Traj-GCN~\cite{TrajGCN}  &35.6 &67.8 &90.6 &106.9 &117.8 \\
      DMGNN~\cite{DMGCN} &37.3 &67.8 &94.5 &109.7 &123.6  \\
      MSR-GCN~\cite{MSRGCN} &37.8 &71.3 &93.9 &110.8 &121.5 \\
      STSGCN~\cite{STSGCN} &37.8 &67.5 &92.8 &106.7 &112.2 \\
      PGBIG \cite{PGBIGGCN} &\underline{29.3} &58.3 &\textbf{79.8} &\underline{94.4} &\underline{104.1} \\
      SPGSN \cite{SPGSNGCN} &32.9 &64.5 &91.6 &104.0 &111.1 \\
      DPnet \cite{DPNet}  &29.5 &\underline{58.0} &84.7 &103.1 &109.3 \\
    GGMotion (Ours)  &\textbf{26.3}  &\textbf{53.8}  &\underline{81.5}  &\textbf{93.9}  &\textbf{103.0} \\
      \hline
      \end{tabular}}
      \vspace{-3mm}
      \label{tab:3DPW}
\end{table}

%% file: tables/mode_size.tex
\begin{table}[t]
    \small 
    \centering
    \renewcommand\arraystretch{1.00}
    \caption{\small Comparisons of MPJPE, running time, model sizes.}
    \vspace{-3mm}
    \resizebox{\linewidth}{!}{
    \begin{tabular}{c|c|c|c} \hline
    Method      & MPJPE (mm) & RunTime (ms) & ModelSize (M)\\ \hline
    Traj-GCN  \cite{TrajGCN}        & 58.42            & \textbf{6.18}            & 2.55         \\ 
    MSRGCN \cite{MSRGCN}        & 57.92            & 32.45            & 6.30          \\ 
    PGBIG \cite{PGBIGGCN} & 54.35            & 112.82            & 1.74         \\ 
    SPGSN \cite{SPGSNGCN} & 54.18            & 512.73            & 5.66         \\ 
    GGMotion (Ours)  & \textbf{52.90}           & 72.85            & \textbf{0.76}      
    \\ \hline
    \end{tabular}
    }
    \label{tab:model_size}
    \vspace{-3mm}
\end{table}

%% file: tables/ablation_module.tex
\begin{table}[t]
  \centering
      \renewcommand\arraystretch{0.9}
    \setlength{\tabcolsep}{2.5pt}
  \caption{\small Ablation study of key modules on Human3.6M.}
    \vspace{-2.6mm}
  \resizebox{1.0\linewidth}{!}{
  \small
    \begin{tabular}{l|cccccc}
    \toprule
     Ablation  & 80ms & 160ms & 320ms & 400ms & 560ms & 1000ms \\
    \midrule
   w/o centroid update & 9.1 & 20.8 & 45.7 & 57.0 & 77.5 & 110.1 \\
   S. field only & 9.1 & 21.0 & 46.1 & 57.7 & 78.1 &110.6 \\
   T. field only & 9.1 & 20.8 & 45.6 & 57.0 &78.2 &110.7\\
   w/o scaling factors & 9.1 & 20.7 & 45.7 & 57.3 & 77.5 & 110.2 \\
   w/o S-T field  & 9.2 & 20.9 & 46.1 & 57.6 & 78.7 & 111.5 \\
   w/o Inter-group & 9.3 & 21.4 & 46.7 & 58.3 & 79.6 & 112.5 \\
   w/o Intra-group &  9.2 &21.1 &46.0 & 57.4 & 77.3 &109.9 \\
   Iterative D-K &  9.4 &21.3&46.4 & 57.9 &78.2 & 110.7\\
   w/o D-K &  9.1 &20.9 &45.7 & 57.2 & 78.0 & 111.2 \\
   Replace MLP & 9.5  &21.7 &47.3 &59.0 &78.8 &111.1 \\
   GGMotion &  \textbf{9.0} &\textbf{20.6}&\textbf{45.1} & \textbf{56.4} & \textbf{76.7} & \textbf{109.6} \\
    \bottomrule
    \end{tabular}
}
  \label{tab:ablation_module}%
  \vspace{-3mm}
\end{table}

%% file: tables/ablation_block.tex
\begin{table}[t]
  \centering
  \renewcommand\arraystretch{0.85}
  \setlength{\tabcolsep}{6pt}
  \caption{\small Ablation study of the number of blocks on Human3.6M.}
    \vspace{-3mm}
  \resizebox{1.0\linewidth}{!}{
  \small
    \begin{tabular}{l|cccc|c}
    \toprule
    Blocks & 80ms & 160ms & 320ms & 400ms &ModelSize (M)\\
    \midrule
   1 & 9.6 & 22.1 & 47.9 &   59.8  &0.12 \\
   2 &  9.2&21.1&46.3 & 57.8  &0.23\\
   3 &  9.1&20.9&45.7 & 57.1 &0.35\\
   4 (Ours) &\textbf{9.0} &\textbf{20.6}&\textbf{45.1} & \textbf{56.4}  &0.46\\
   5 &  9.1&20.8&45.6 & 57.2  &0.58\\
   \bottomrule
   \end{tabular}
}
  \label{tab:ablation_block}%
  \vspace{-3mm}
\end{table}

%% file: tables/ablation_group.tex
\begin{table}[t]
  \centering
        \renewcommand\arraystretch{0.85}
    \setlength{\tabcolsep}{4.5pt}
  \caption{\small Ablation study of body groups on Human3.6M.}
    \vspace{-3mm}
  \resizebox{1.0\linewidth}{!}{
  \small
    \begin{tabular}{l|cccccc}
    \toprule
     Body groups  & 80ms & 160ms & 320ms & 400ms & 560ms & 1000ms\\
    \midrule
    1 &9.4  &21.2  &46.7 & 58.3 &79.1 &111.2 \\
    2 &9.1  &20.8 &46.0 &57.6 &77.9 &110.1 \\
    5 &9.1 &20.7 &45.3 & 56.7 & 77.2 &110.1\\
    6 (Ours) &\textbf{9.0} &\textbf{20.6}&\textbf{45.1} & \textbf{56.4} & \textbf{76.7} & \textbf{109.6} \\
    \bottomrule
    \end{tabular}
}
  \label{tab:ablation_group}%
  \vspace{-3mm}
\end{table}

%% file: tables/ablation_loss.tex
\begin{table}[!htbp]
  \centering
        \renewcommand\arraystretch{0.85}
    \setlength{\tabcolsep}{4.5pt}
  \caption{\small Ablation study of auxiliary loss terms on Human3.6M.}
    \vspace{-3mm}
  \resizebox{\linewidth}{!}{
  \footnotesize
    \begin{tabular}{cc|ccccccc}
    \toprule
        \(\mathcal{L}_{\text{pos}}\) & \(\mathcal{L}_{\text{aux}}\)  & 80ms & 160ms & 320ms & 400ms & 560ms & 1000ms  \\
    \midrule 
   \checkmark & &9.3 & 21.0 & 45.8 & 57.3 & 77.7 & 110.5  \\
      \checkmark&\checkmark &\textbf{9.0} &\textbf{20.6}&\textbf{45.1} & \textbf{56.4} & \textbf{76.7} & \textbf{109.6} \\
    \bottomrule
    \end{tabular}
    }
  \label{tab:ablation_loss}%
\vspace{-3mm}
\end{table}